\documentclass{article}

\usepackage[preprint]{neurips_2023}

\usepackage[utf8]{inputenc} % allow utf-8 input
\usepackage[T1]{fontenc}    % use 8-bit T1 fonts
\usepackage{hyperref}       % hyperlinks
\usepackage{url}            % simple URL typesetting
\usepackage{booktabs}       % professional-quality tables
\usepackage{amsfonts}       % blackboard math symbols
\usepackage{nicefrac}       % compact symbols for 1/2, etc.
\usepackage{microtype}      % microtypography
\usepackage{xcolor}         % colors
\usepackage{graphicx}  
\usepackage{subcaption}  
\usepackage{bm}

\title{CLIP-Mamba: CLIP Pretrained Mamba Models with OOD and Hessian Evaluation}

\author{%
  Weiquan Huang, Yifei Shen, Yifan Yang \thanks{Weiquan Huang is with Tongji Univeristy(weiquanh@tongji.edu.cn). Yifan Yang, Yifei Shen are with Microsoft.}
}

\begin{document}

\maketitle

\begin{abstract}
  State space models and Mamba-based models have been increasingly applied across various domains, achieving state-of-the-art performance. This technical report introduces the first attempt to train a transferable Mamba model utilizing contrastive language-image pretraining (CLIP). We have trained Mamba models of varying sizes and undertaken comprehensive evaluations of these models on $26$ zero-shot classification datasets and $16$ out-of-distribution (OOD) datasets. Our findings reveal that a Mamba model with 67 million parameters is on par with a 307 million-parameter Vision Transformer (ViT) model in zero-shot classification tasks, highlighting the parameter efficiency of Mamba models. In tests of OOD generalization, Mamba-based models exhibit exceptional performance in conditions of OOD image contrast or when subjected to high-pass filtering. However, a Hessian analysis indicates that Mamba models feature a sharper and more non-convex landscape compared to ViT-based models, making them more challenging to train. The source code is available at \url{https://github.com/raytrun/mamba-clip}.
\end{abstract}

\section{Introduction}
Foundation models, i.e., models pretrained on massive data and adapted for specific downstream tasks, have emerged as a vibrant field within machine learning. The transformative six years preceding have seen Transformers establish themselves as the principal architecture underpinning foundation models across a multitude of domains \cite{dosovitskiy2020image,vaswani2017attention,ying2021transformers,gong2021ast,zhou2021informer,cui2024scgpt}. The core of the Transformer architecture is the self-attention mechanism, which intricately facilitates the flow of information between every token pair. This mechanism is critically acclaimed for its indispensable role in facilitating in-context learning \cite{wen2024rnns}, enhancing reasoning capabilities \cite{yang2024efficient}, and bolstering out-of-distribution (OOD) robustness \cite{li2022sparse}. Nonetheless, the self-attention mechanism's quadratic computational demands pose significant scalability challenges, particularly concerning window length, thereby emerging as a substantial impediment for practical applications. In response, a wealth of research has been dedicated to devising efficient self-attention mechanisms capable of operating within sub-quadratic time \cite{wang2020linformer,katharopoulos2020transformers,choromanski2020rethinking}. Despite these advancements, such innovations often demonstrate inferior performance when compared with their quadratic-time Transformer counterparts.

Selective state space models (Mamba) \cite{gu2023mamba} have recently emerged as promising candidates for the next-generation foundation model backbone as they exhibit better scaling laws than Transformers while enjoying linear-time complexity. In the brief span of the last few months, the Mamba model has demonstrated remarkable success across a spectrum of critical domains, including but not limited to, natural language processing \cite{gu2023mamba,qiao2024vl}, image processing \cite{zhu2024vision,liu2024vmamba}, video analysis \cite{yang2024vivim,li2024videomamba}, time-series forecasting \cite{tang2024vmrnn,patro2024simba}, graph theory applications \cite{wang2024graph,behrouz2024graph}, point cloud processing \cite{liang2024pointmamba,zhang2024point}, recommendation systems \cite{yang2024uncovering}, reinforcement learning \cite{rimon2024mamba}, and medical diagnostics \cite{ma2024u,xing2024segmamba}. Focusing on computer vision, a myriad of Mamba-based models have emerged, setting new state-of-the-art baselines in image classification \cite{zhu2024vision,patro2024simba}, object detection \cite{liu2024vmamba}, segmentation \cite{liu2024vmamba,ma2024u}, image restoration \cite{zheng2024u,guo2024mambair}, and 3D reconstruction \cite{shen2024gamba}. Despite these achievements, current Mamba-based models are trained on a fixed array of predetermined object categories, and lacks of zero-shot generalization capabilities. Bridging this gap necessitates the integration of large-scale language-image pretraining, and this is an indispensable component for the evolution of Mamba-based foundational models.

This technical report presents the first attempt to train Mamba models with contrastive language-image pretraining. Specifically, the conclusions of this technical report summarized as follows:
\begin{itemize}
    \item \textbf{CLIP-Mamba models:}  We release the open-sourced CLIP-Mamba models. A Mamba model with 50 million parameters surpasses the performance of an 84 million-parameter ViT model, and a 67 million-parameter Mamba model equates to the performance of a 307 million-parameter ViT model on $26$ zero-shot classification datasets. These results underscore the efficiency and effectiveness of Mamba models.  
    
    \item \textbf{OOD generalization evaluation:} Our extensive evaluations on $16$ OOD datasets demonstrate that Mamba models consistently outperform ViT models. Mamba-based models show exceptional robustness in conditions of OOD image contrast or when subjected to high-pass filtering.

    \item \textbf{Landscape evaluation:} Through the visualization of the Hessian, we delve into the training landscape of Mamba models. Our findings indicate that Mamba models exhibit a more "non-convex" and sharper landscape compared to ViT models, suggesting greater challenges in optimization.
\end{itemize}

\section{Experiments and Analysis}
In this section, we conduct comprehensive experiments and analysis for the CLIP Mamba models versus CLIP Vision Transformer models, in terms of zero-shot classification, OOD generalization, and Hessian spectra. 

\subsection{Zero-shot Classification}
In our study, we train a series of models including VMamba-30M, VMamba-50M, VMamba-89M \cite{liu2024vmamba}, and Simba-L 66.6M \cite{patro2024simba}, utilizing the standard CLIP pretraining pipelines. The zero-shot performance of these models is systematically evaluated across a variety of datasets and summarized in Table \ref{tab:CLIP}. Notably, the 50M-parameter Mamba-S model demonstrates superior performance over the 84M-parameter ViT-B model in the majority of the datasets examined. When considering the pinnacle of performance, the results are evenly split; the 66.6M-parameter Simba-L leads in half of the datasets, while 307M-parameter ViT-L dominates in the remaining half.

\begin{table}[]
\centering
\setlength{\aboverulesep}{0pt}
\setlength{\belowrulesep}{0pt}
\setlength{\tabcolsep}{2pt}
\linespread{1}
\scriptsize
\scalebox{0.9}{
\begin{tabular}{c|cccccccccccccccccccccccccc}
% \hline
% Models       & Food-101 & CIFAR-10 & CIFAR-100 & CUB  & SUN397 & Cars & Aircraft & DTD  & Pets & Caltech-101 & Flowers & MNIST & FER-2013 & STL-10 & EuroSAT & RESISC45 & GTSRB & KITTI & Country211 & PCAM & UFC101 & Kinetics700 & CLVER & HatefulMemes & SST2 & ImageNet & Average \\ \hline
\toprule
  %\rotatebox[origin=lb]{90}{\smash{Models}} &
  Models&
  \rotatebox[origin=lb]{90}{\smash{Food-101}} & \rotatebox[origin=lb]{90}{\smash{CIFAR-10}} & \rotatebox[origin=lb]{90}{\smash{CIFAR-100}} & \rotatebox[origin=lb]{90}{\smash{CUB}} & \rotatebox[origin=lb]{90}{\smash{SUN397}} &
\rotatebox[origin=lb]{90}{\smash{Cars}} & \rotatebox[origin=lb]{90}{\smash{Aircraft}} & \rotatebox[origin=lb]{90}{\smash{DTD}} & \rotatebox[origin=lb]{90}{\smash{Pets}} & \rotatebox[origin=lb]{90}{\smash{Caltech-101}} &
\rotatebox[origin=lb]{90}{\smash{Flowers}} & \rotatebox[origin=lb]{90}{\smash{MNIST}} & \rotatebox[origin=lb]{90}{\smash{FER-2013}} & \rotatebox[origin=lb]{90}{\smash{STL-10}} & \rotatebox[origin=lb]{90}{\smash{EuroSAT}} &
\rotatebox[origin=lb]{90}{\smash{RESISC45}} & \rotatebox[origin=lb]{90}{\smash{GTSRB}} & \rotatebox[origin=lb]{90}{\smash{KITTI}} & \rotatebox[origin=lb]{90}{\smash{Country211}} & \rotatebox[origin=lb]{90}{\smash{PCAM}} &
\rotatebox[origin=lb]{90}{\smash{UCF101}} & \rotatebox[origin=lb]{90}{\smash{Kinetics700}} & \rotatebox[origin=lb]{90}{\smash{CLEVR}} & \rotatebox[origin=lb]{90}{\smash{HatefulMemes}} & \rotatebox[origin=lb]{90}{\smash{SST2}} &
\rotatebox[origin=lb]{90}{\smash{ImageNet}}\\ \hline %& \rotatebox[origin=lb]{90}{\smash{Average}} 
VMamba\_B (89M)    & 48.5     & 58.0     & 29.9      & 36.5 & 50.4   & 5.8  & 8.5      & 26.5 & 30.2 & 64.7        & 52.8    & 9.7   & 19.6     & 91.9   & 16.0    & 30.4     & 7.9   & 40.2  & 10.2       & {\color {red} 59.9} & 35.2   & 25.6        & 12.6  & 51.6         & 50.1 & 38.3         \\ \hline %& 35.0
VMamba\_S (50M)   & 49.4     & 70.3     & 34.3      & 39.1 & 53.9   & {\color {red} 6.9}  & 8.4      & 26.0 & 31.3 & 68.7        & 54.1    & 10.1  & 9.8      & 92.8   & 17.6    & 31.4     & 6.9   & 23.5  & 10.9       & 54.2 & {\color {red} 38.4}   & 27.1        & {\color {red} 13.2}  & 50.5         & 50.0 & 40.0         \\ \hline %& 35.3
VMamba\_T220 (30M) & 46.5     & 50.9     & 22.9      & 35.6 & 51.1   & 5.7  & 6.8      & 25.1 & 31.0 & 64.9        & 54.0    & 10.1  & 12.5     & 91.6   & 13.9    & 25.4     & {\color {red} 10.7}  & 32.3  & 9.9        & 55.0 & 34.0   & 25.1        & 12.7  & 53.9         & 50.6 & 38.7         \\ \hline %& 33.5
Simba\_L (66.6M)     & 52.7     & 67.4     & 31.0      & 39.1 & 52.7   & {\color {red} 6.9}  & {\color {red} 9.1}      & {\color {red} 27.8} & {\color {red} 33.4} & {\color {red} 68.9}        & {\color {red} 55.9}    & 8.0   & 16.0     & 93.9   & 17.4    & {\color {red} 32.3}     & 8.9   & {\color {red} 41.5}  & 11.1       & 58.1 & 35.7  & {\color {red} 27.9}        & 12.1  & {\color {red} 54.9}         & 50.1 & {\color {red} 41.6}         \\ \hline %& 36.7
VIT\_B(84M)       & 50.6     & 66       & 34.5      & 38.8 & 51.1   & 4.0    & 5.4      & 21.2 & 28.5 & 60.9        & 53.3    & 8.4   & 17.3     & 90.5   & {\color {red} 30.2}    & 21.5     & 6.1   & 35.1  & 10.5       & 53.5 & 28.5   & 22.1        & 10.8  & 52.4         & {\color {red} 50.7} & 37.6         \\ \hline %& 34.2
VIT\_L(307M)       & {\color {red}59.5}     & {\color {red} 72.9}     & {\color {red} 41.5}      & {\color {red} 40.3} & {\color {red} 53.6}   & {\color {red} 6.9}  & 6.4      & 20.6 & 27.9 & 65.4        & 55      & {\color {red} 10.3}  & {\color {red} 34.5}     & {\color {red} 94.2}   & 22.7    & 28.8     & 5.8   & 41.4  & {\color {red} 12.5}       & 54.9 & 34.3   & 24.0          & 12.9  & 54.3         & 50.1 & 40.4         \\ \hline %& {\color {red} 37.4}
\end{tabular}
}
\caption{Zero-shot performance of different architectures trained with CLIP.}
\label{tab:CLIP}
\end{table}

\subsection{OOD Robustness and Comparison with Humans}

\begin{figure}[t]  
\centering  
% Row 1  
\begin{subfigure}{0.6\textwidth}  
    \includegraphics[width=\linewidth]{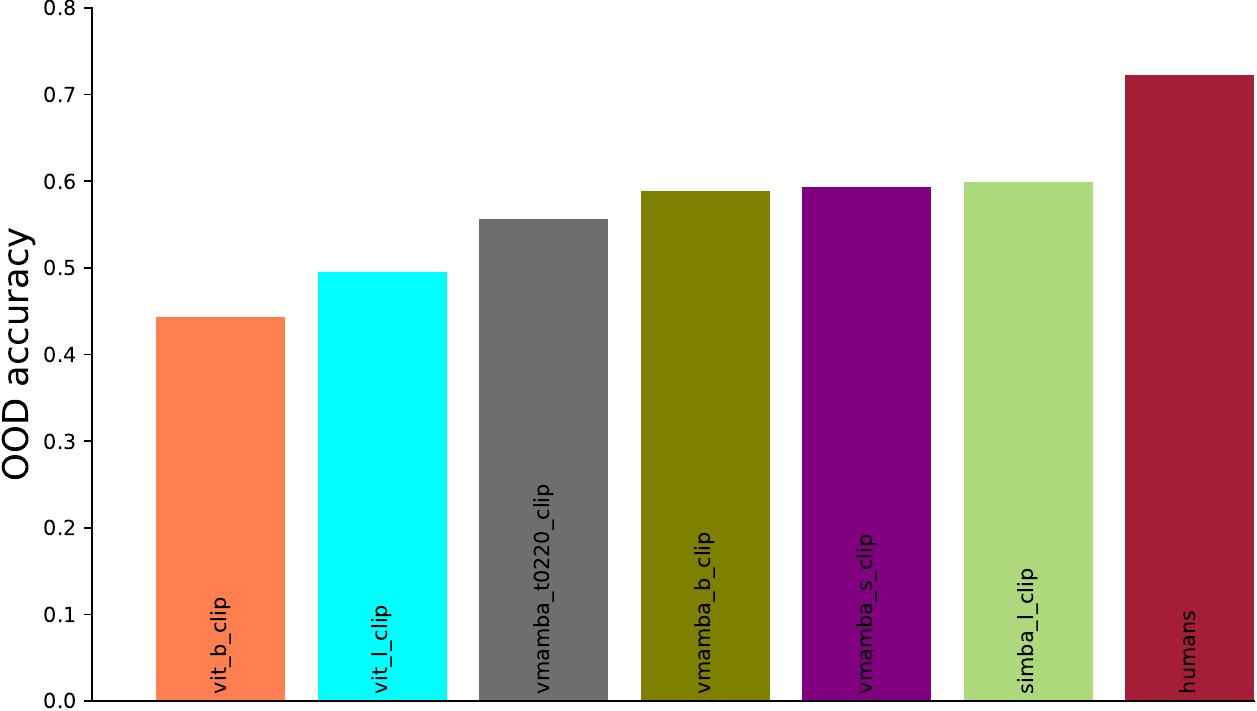}  
    \caption{Overall OOD accuracy.}   
\end{subfigure}  
\hfill  
\begin{subfigure}{0.35\textwidth}  
    \includegraphics[width=\linewidth]{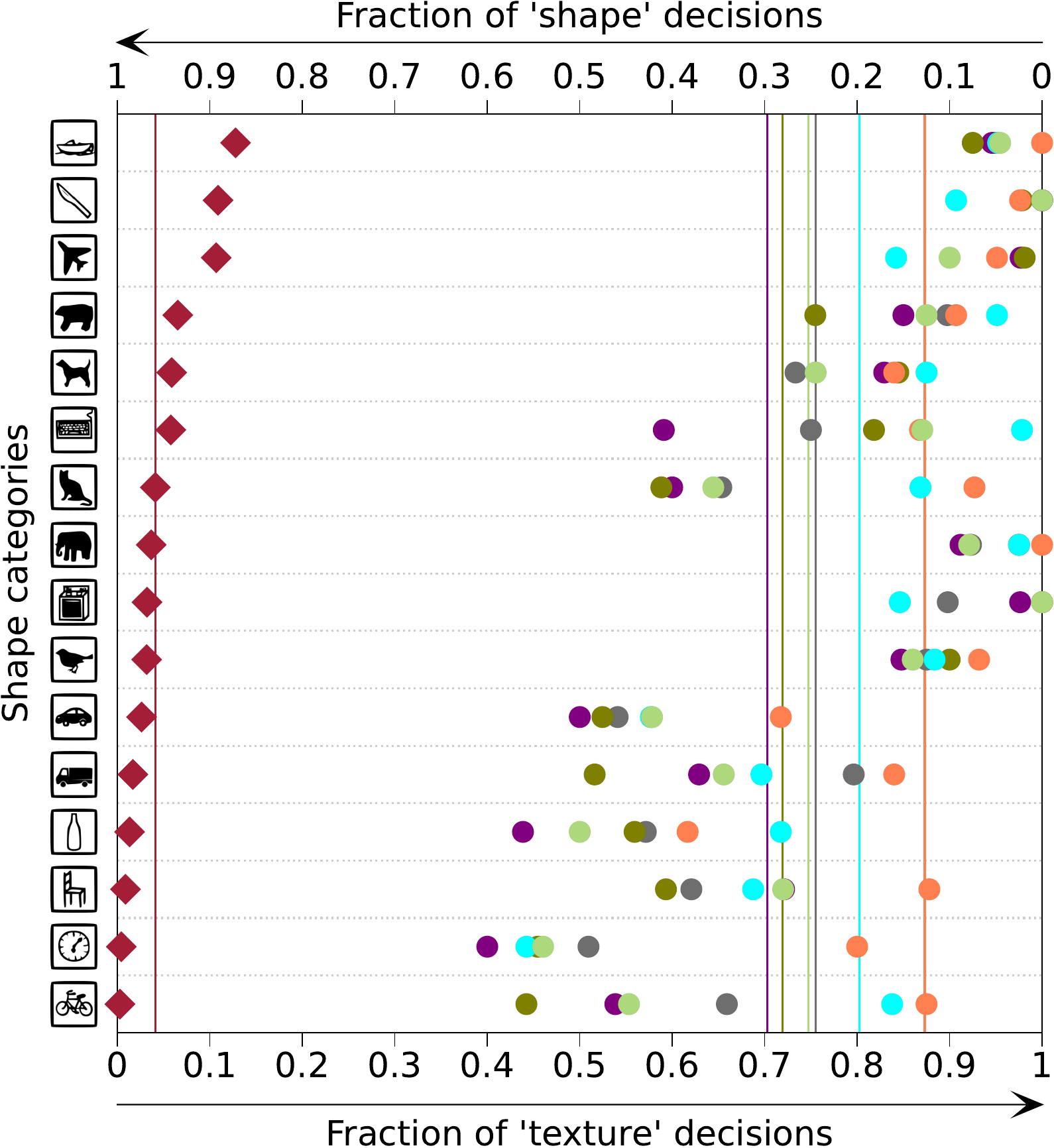}  
    \caption{Shape bias decisions.}  
\end{subfigure}  
\caption{Overall performance and shape bias.}  
\label{fig:OOD}  
\end{figure}  

\begin{figure}[!h]  
\centering  
% Row 1  
\begin{subfigure}{0.24\textwidth}  
    \includegraphics[width=\linewidth]{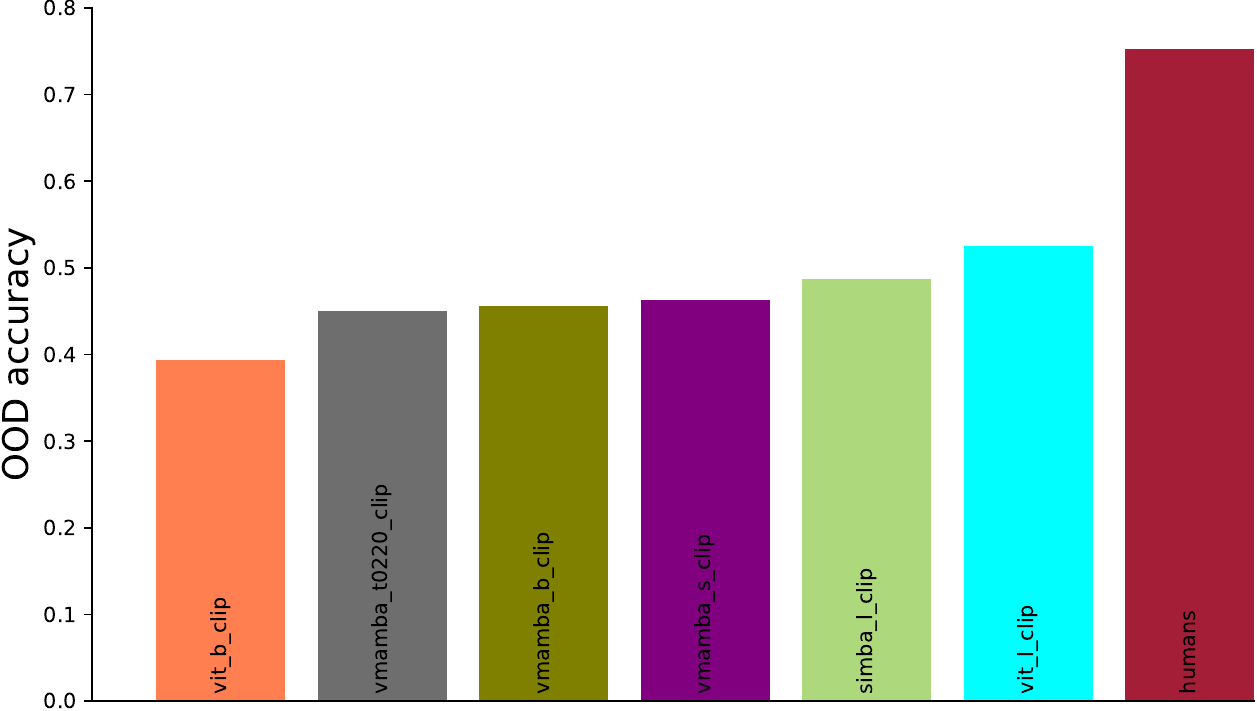}  
    \caption{Silhouette}   
\end{subfigure}  
\hfill  
\begin{subfigure}{0.24\textwidth}  
    \includegraphics[width=\linewidth]{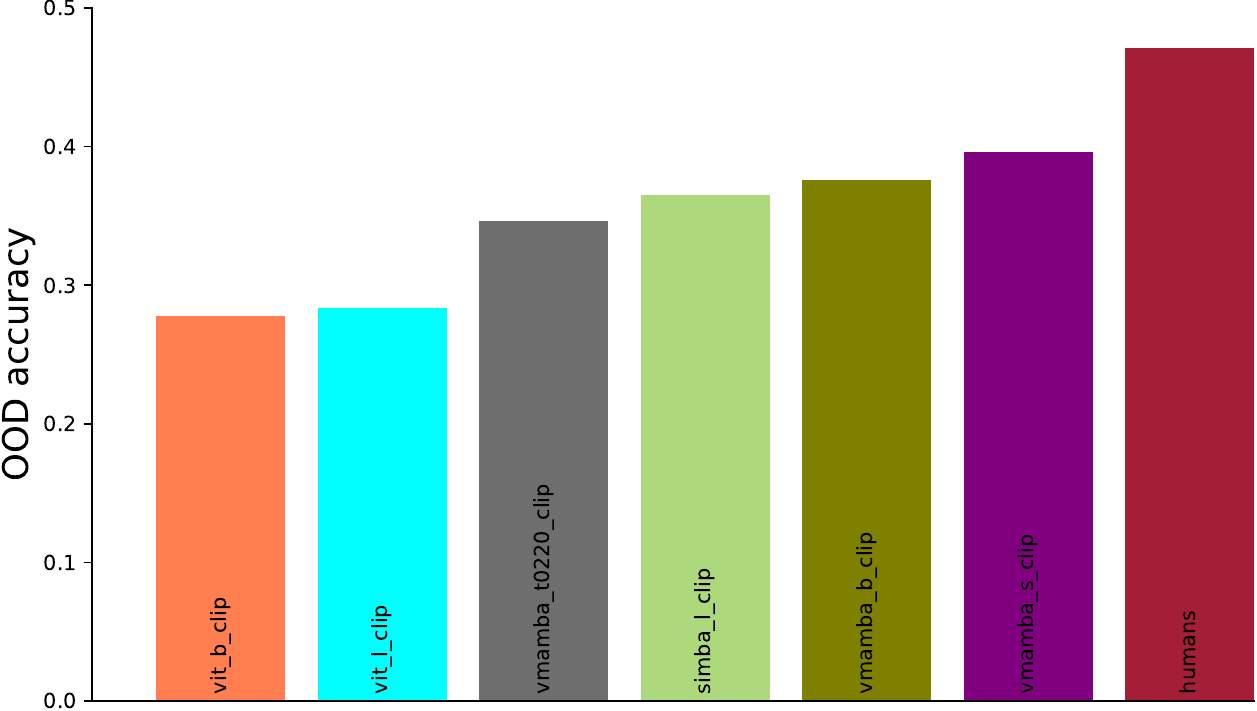}  
    \caption{Stylized}  
\end{subfigure}  
\hfill  
\begin{subfigure}{0.24\textwidth}  
    \includegraphics[width=\linewidth]{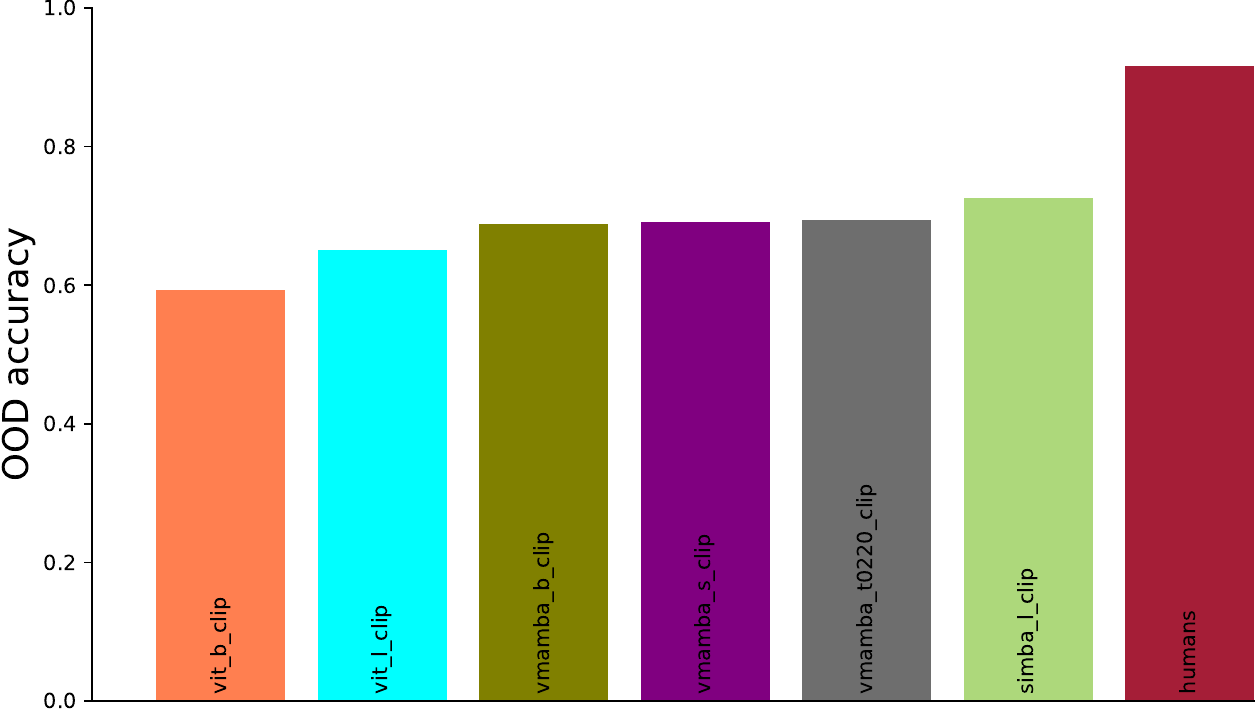}  
    \caption{Sketch}   
\end{subfigure}  
\hfill  
\begin{subfigure}{0.24\textwidth}  
    \includegraphics[width=\linewidth]{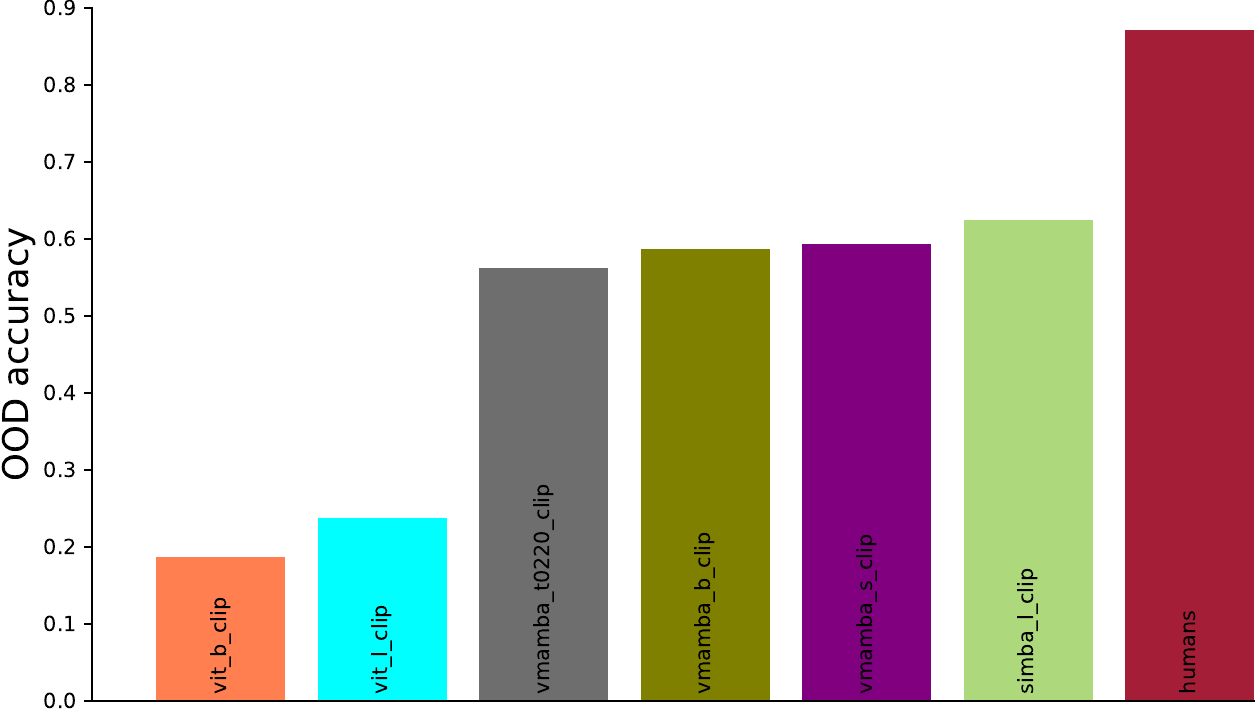}  
    \caption{Edge}  
\end{subfigure}  

\begin{subfigure}{0.24\textwidth}  
    \includegraphics[width=\linewidth]{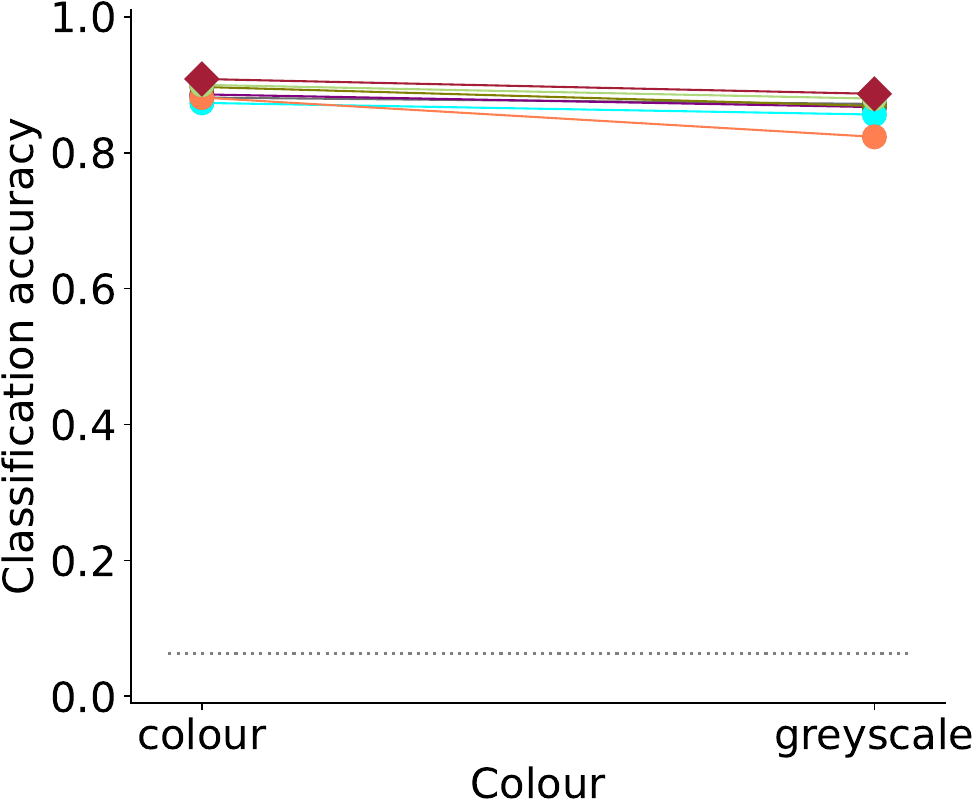}  
    \caption{Color}   
\end{subfigure}  
\hfill  
\begin{subfigure}{0.24\textwidth}  
    \includegraphics[width=\linewidth]{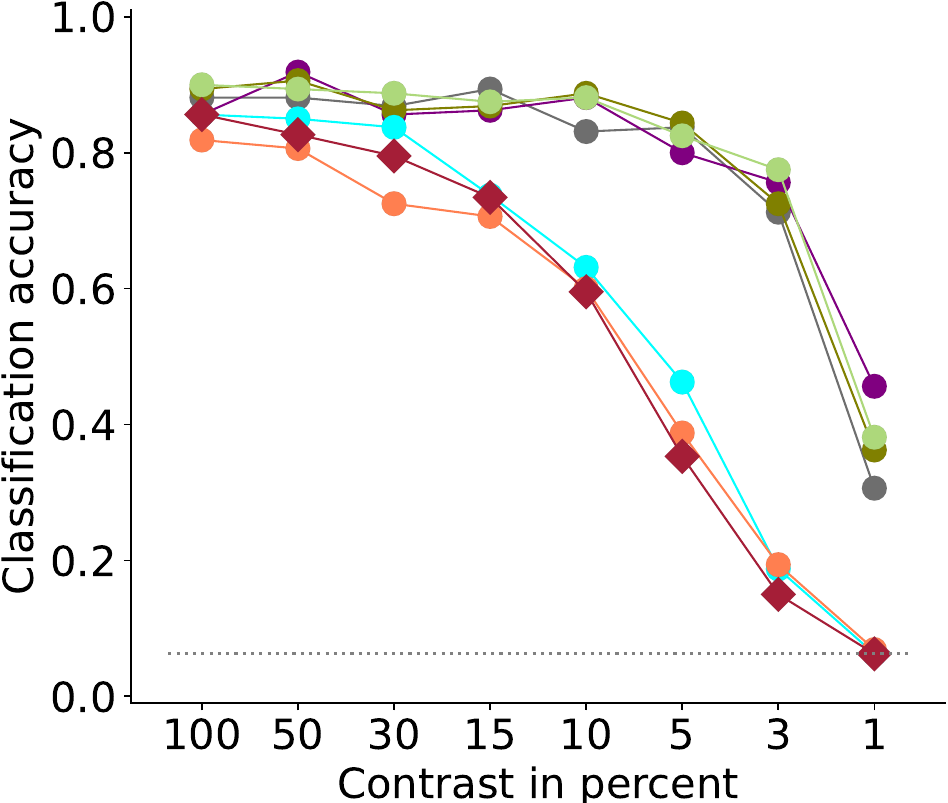}  
    \caption{Contrast}   
\end{subfigure}  
\hfill  
\begin{subfigure}{0.24\textwidth}  
    \includegraphics[width=\linewidth]{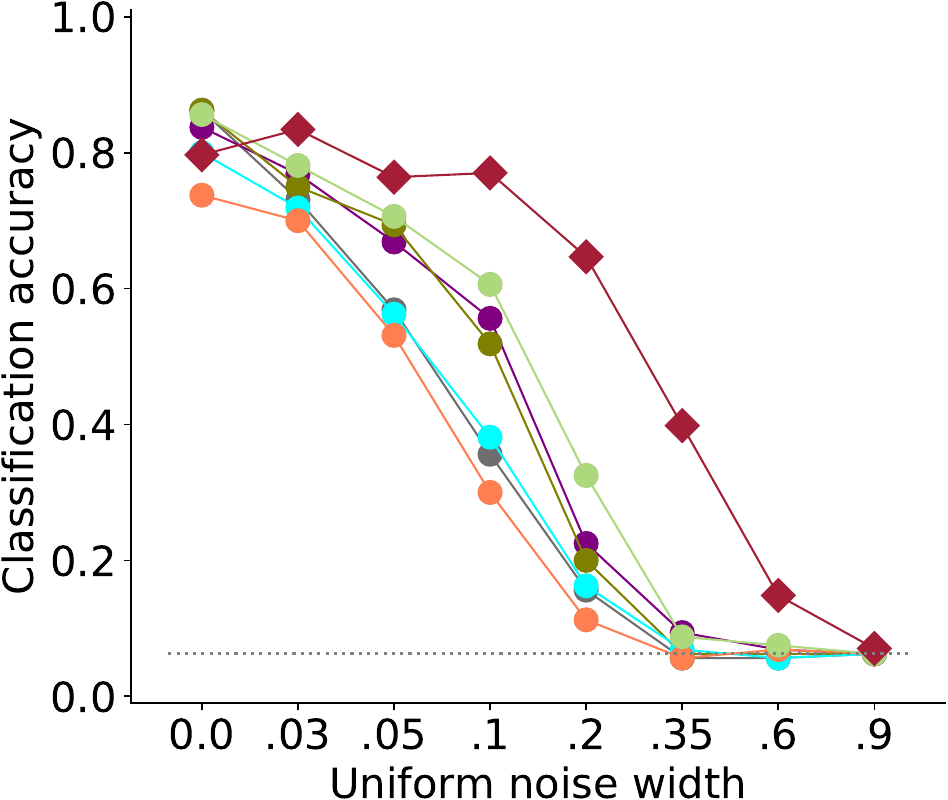}  
    \caption{Uniform noise}  
\end{subfigure}  
\hfill  
\begin{subfigure}{0.24\textwidth}  
    \includegraphics[width=\linewidth]{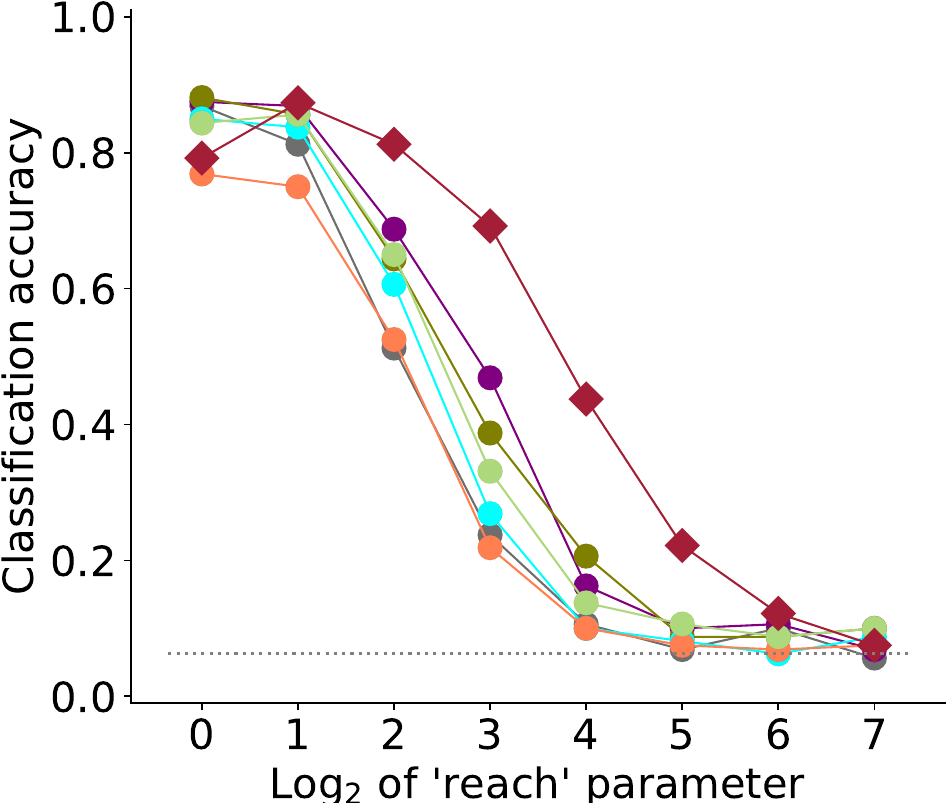}  
    \caption{Eidolon I}  
\end{subfigure}  

\begin{subfigure}{0.24\textwidth}  
    \includegraphics[width=\linewidth]{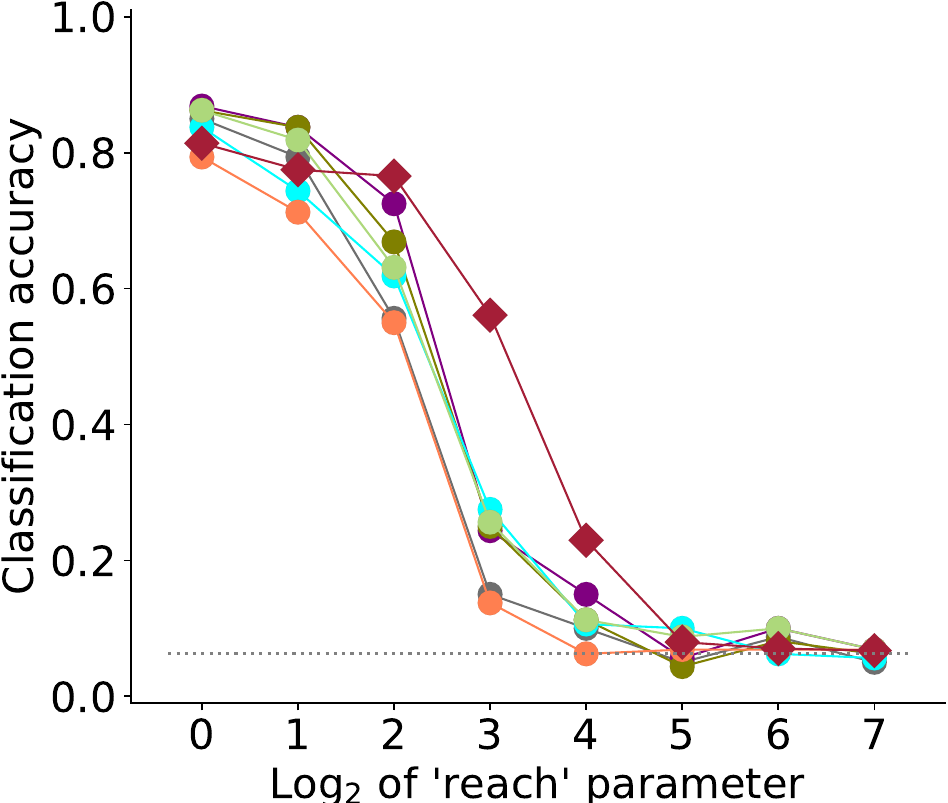}  
    \caption{Eidolon II}   
\end{subfigure}  
\hfill  
\begin{subfigure}{0.24\textwidth}  
    \includegraphics[width=\linewidth]{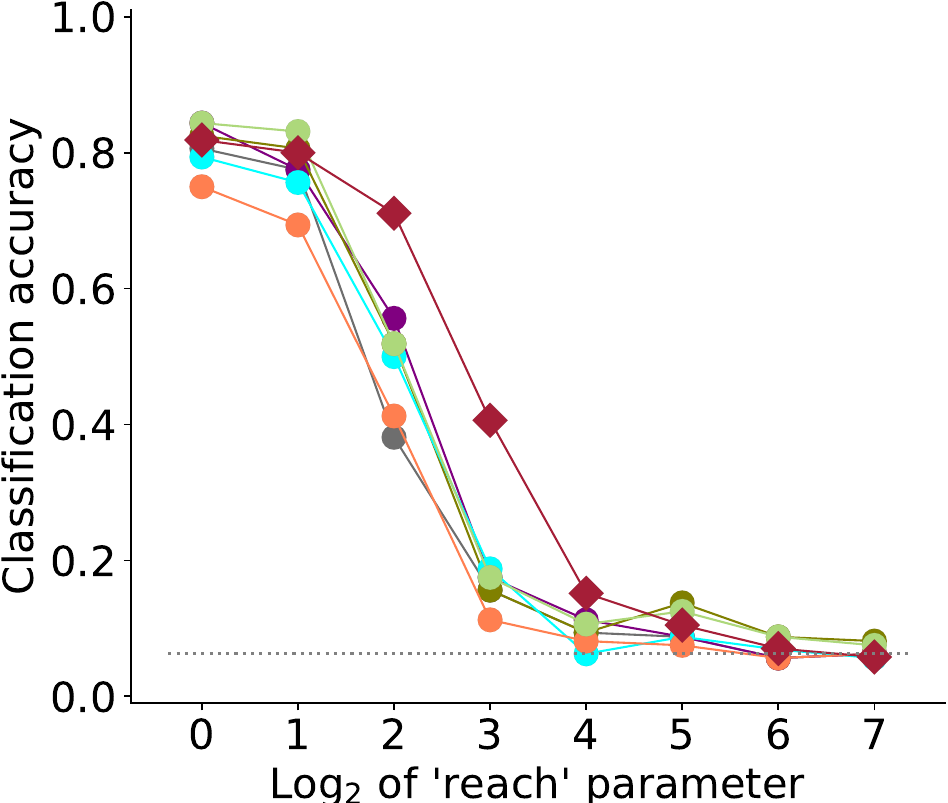}  
    \caption{Eidolon III}   
\end{subfigure}  
\hfill  
\begin{subfigure}{0.24\textwidth}  
    \includegraphics[width=\linewidth]{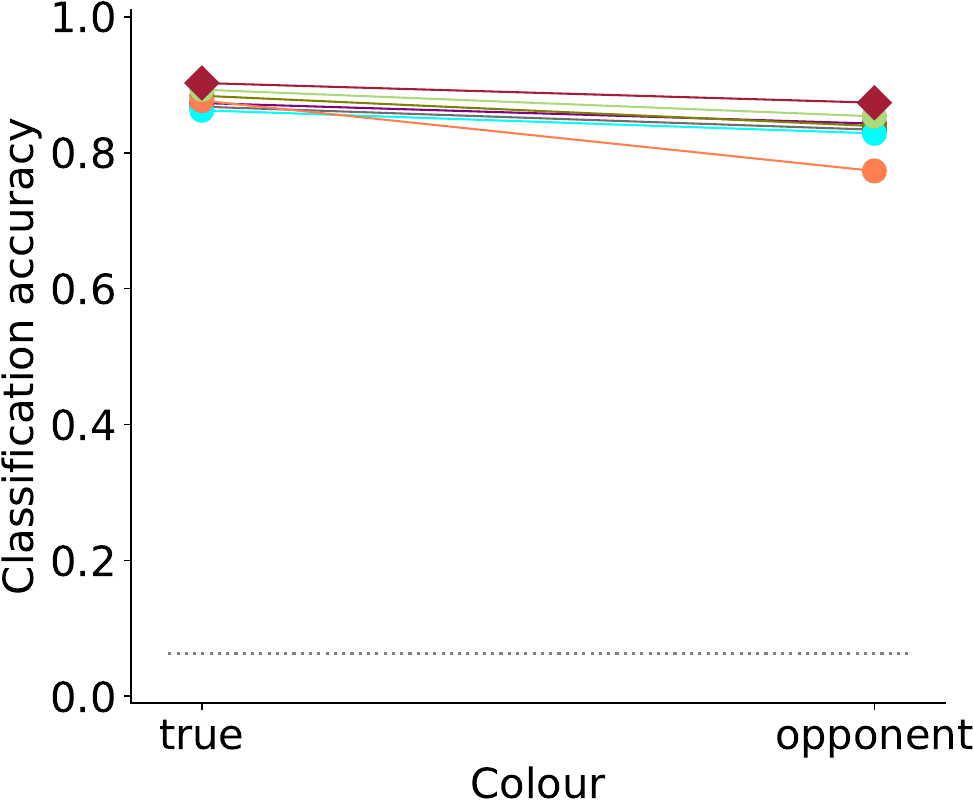}  
    \caption{False Color}  
\end{subfigure}  
\hfill  
\begin{subfigure}{0.24\textwidth}  
    \includegraphics[width=\linewidth]{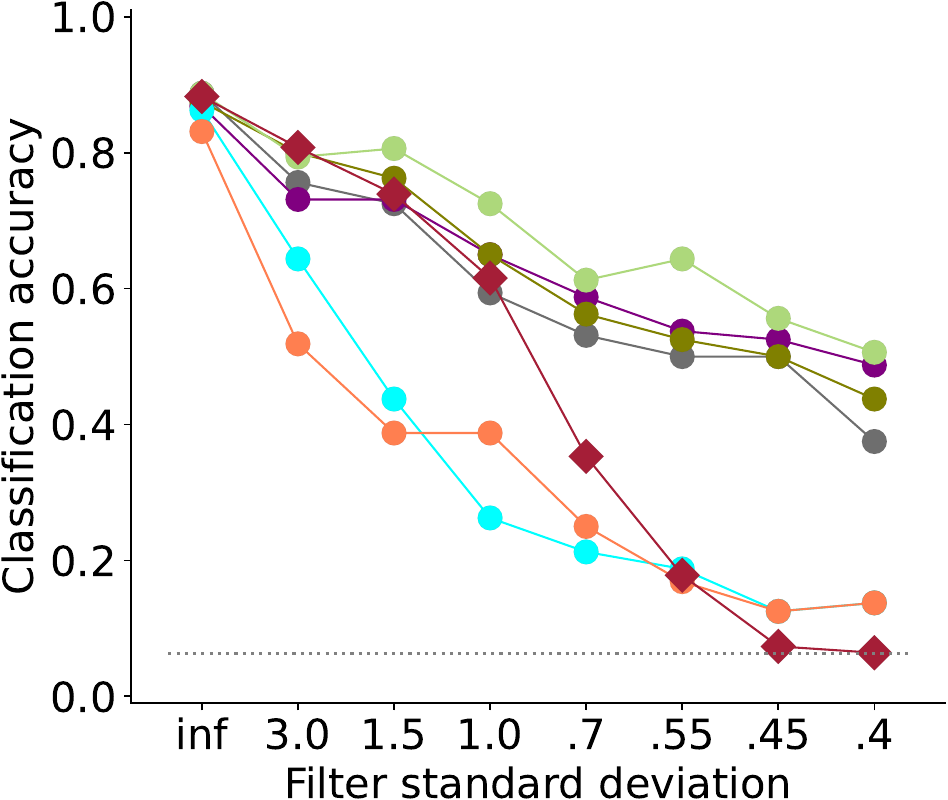}  
    \caption{High-pass}  
\end{subfigure}

\begin{subfigure}{0.24\textwidth}  
    \includegraphics[width=\linewidth]{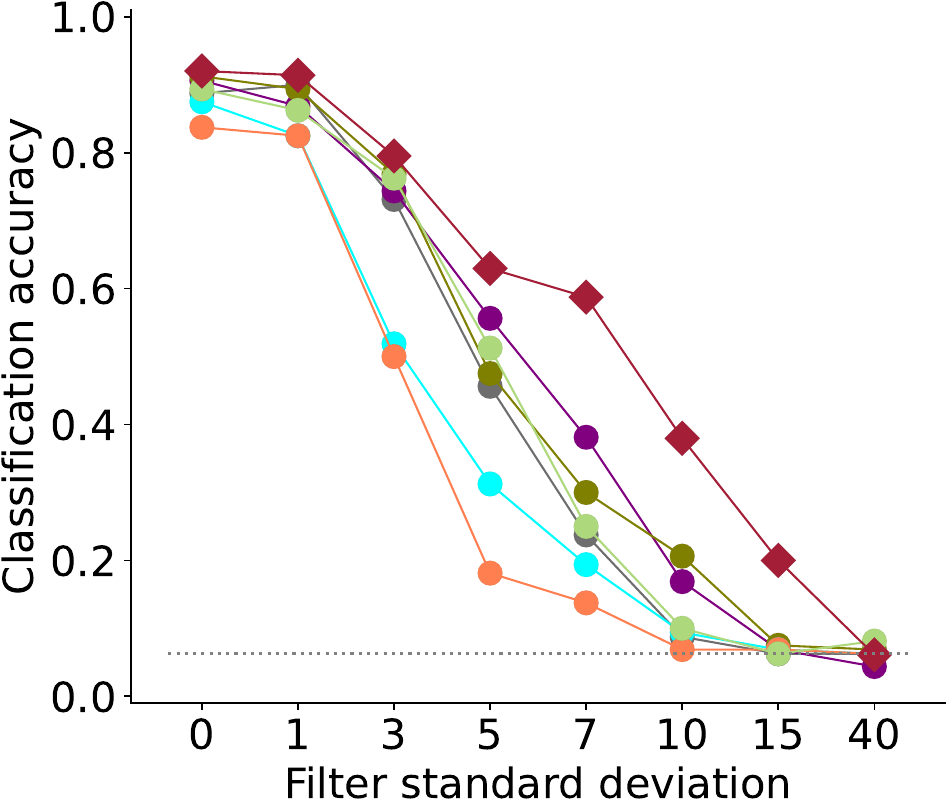}  
    \caption{Low-pass}   
\end{subfigure}  
\hfill  
\begin{subfigure}{0.24\textwidth}  
    \includegraphics[width=\linewidth]{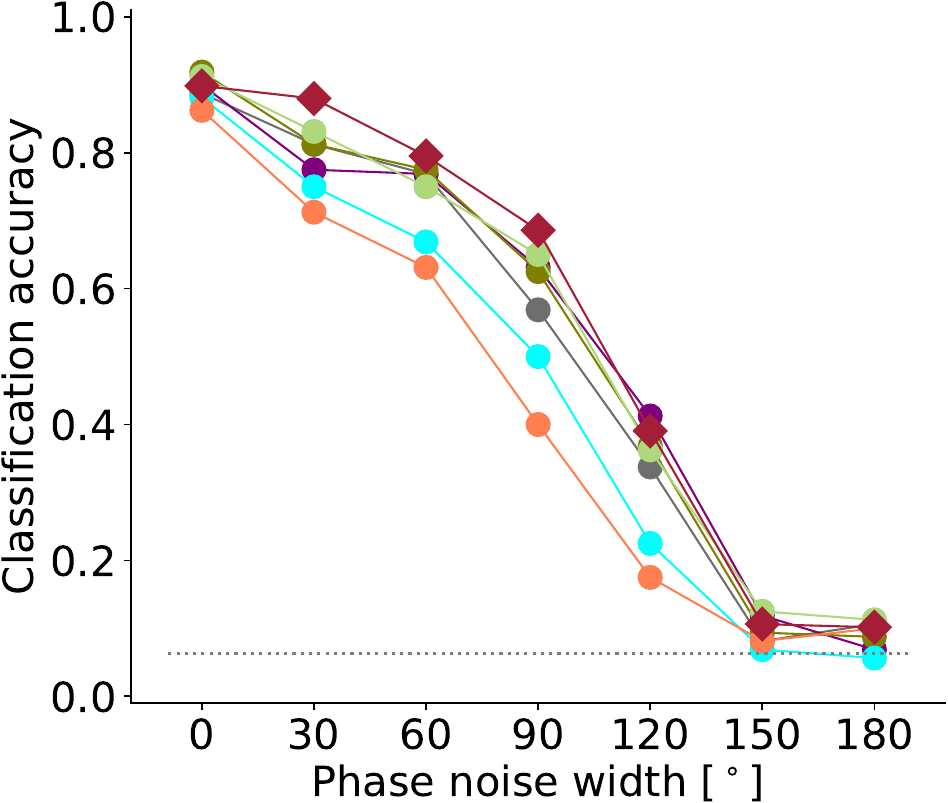}  
    \caption{Phase Scrambling}   
\end{subfigure}  
\hfill  
\begin{subfigure}{0.24\textwidth}  
    \includegraphics[width=\linewidth]{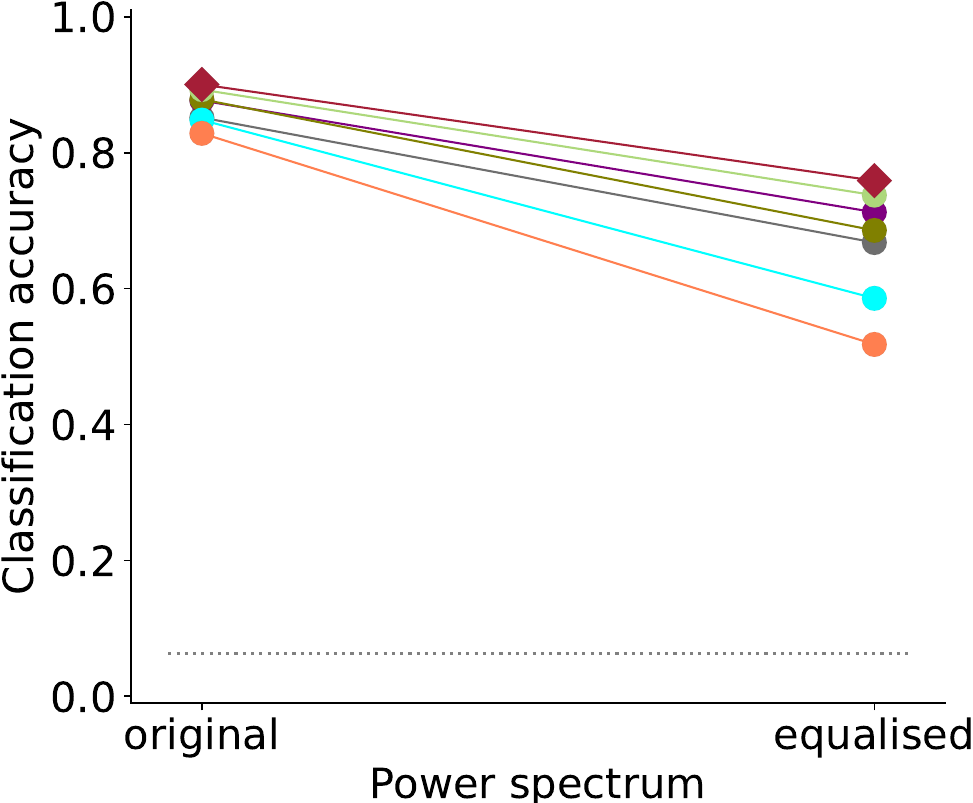}  
    \caption{Power Equalization}  
\end{subfigure}  
\hfill  
\begin{subfigure}{0.24\textwidth}  
    \includegraphics[width=\linewidth]{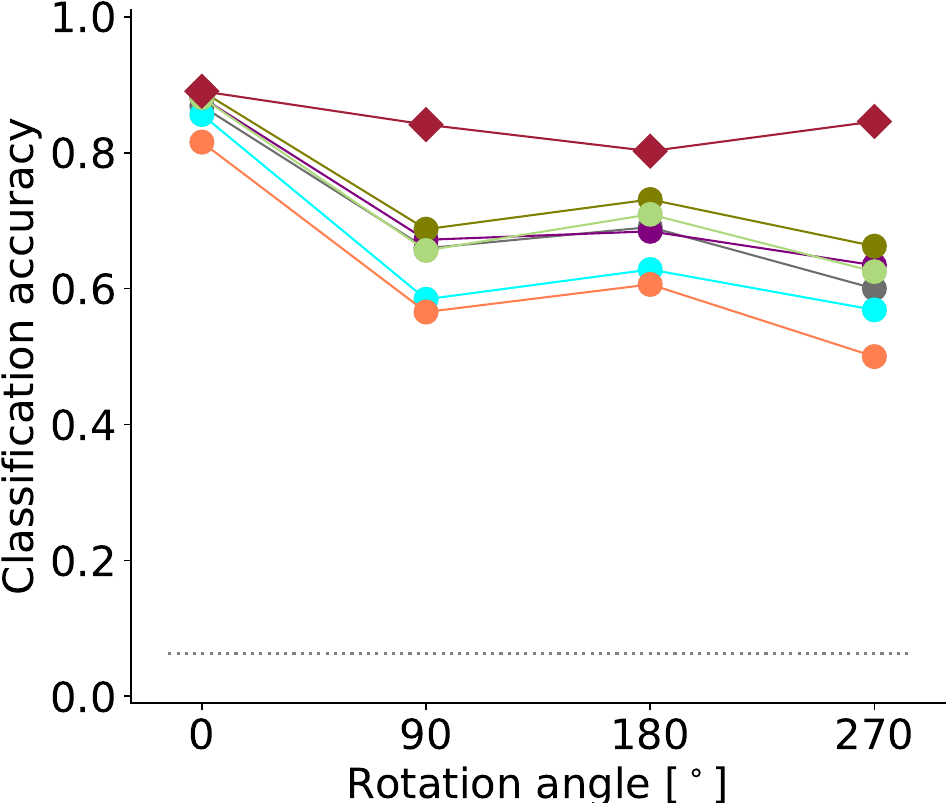}  
    \caption{Rotation}  
\end{subfigure}  

\caption{Detailed performance on 16 OOD datasets.}  
\label{fig:OOD_details}  
\end{figure}

Building upon the methodology outlined by \cite{geirhos2021partial}, we delve into a comparative analysis involving VMamba, Simba, ViTs (Vision Transformers), and human performance across 16 Out-of-Distribution (OOD) datasets. The results of this comprehensive comparison are visually represented in Figure \ref{fig:OOD}, which provides an overview of the overall performance, and in Figure \ref{fig:OOD_details}, which offers a detailed breakdown of performance metrics.

\begin{figure}[t]  
\centering  
% Row 1  
\begin{subfigure}{0.30\textwidth}  
    \includegraphics[width=\linewidth]{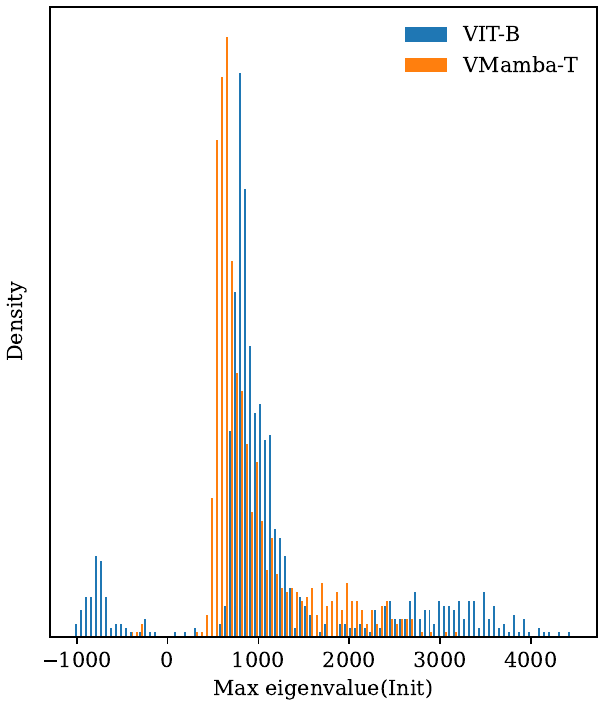}  
    \caption{At initialization.}   
\end{subfigure}  
\hfill  
\begin{subfigure}{0.30\textwidth}  
    \includegraphics[width=\linewidth]{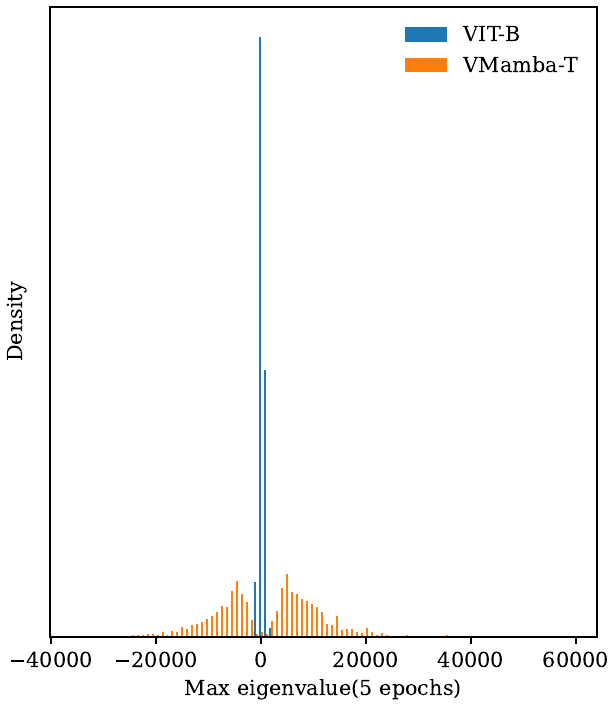}  
    \caption{$5$-th epoch.}  
\end{subfigure}  
\hfill  
\begin{subfigure}{0.30\textwidth}  
    \includegraphics[width=\linewidth]{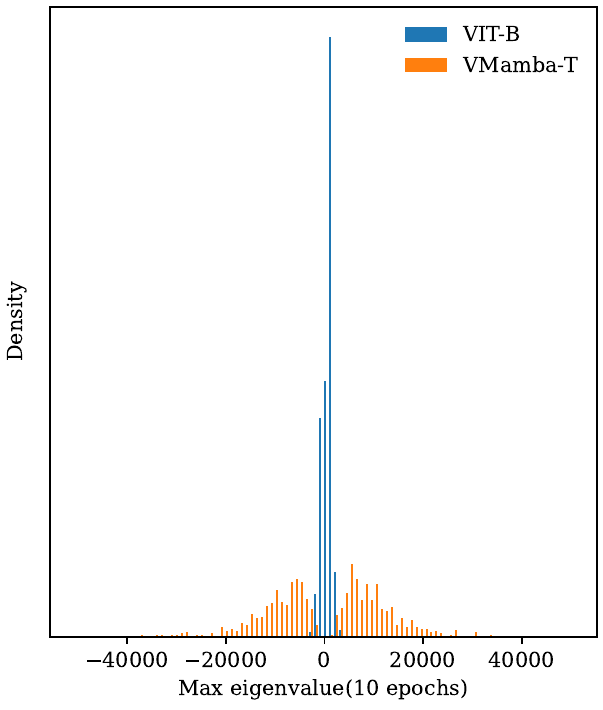}  
    \caption{$10$-th epoch.}  
\end{subfigure}  

\begin{subfigure}{0.30\textwidth}  
    \includegraphics[width=\linewidth]{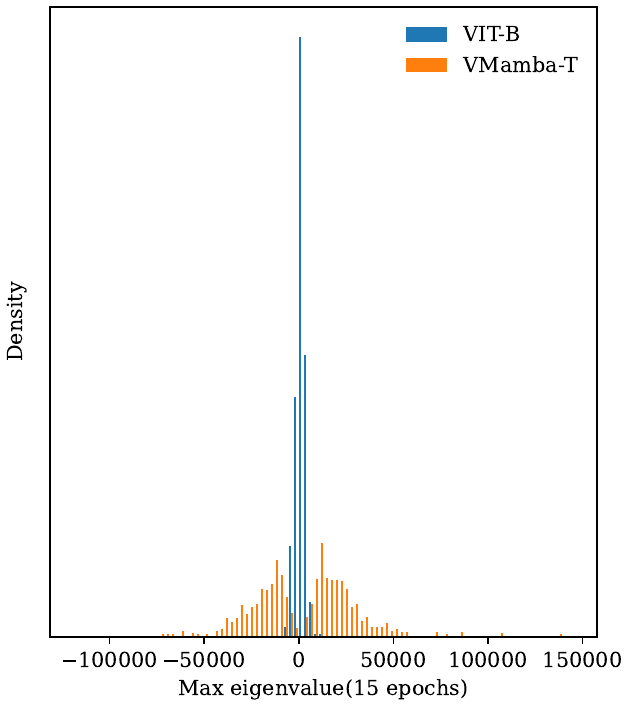}  
    \caption{$15$-th epoch.}  
\end{subfigure}  
\hfill  
\begin{subfigure}{0.30\textwidth}  
    \includegraphics[width=\linewidth]{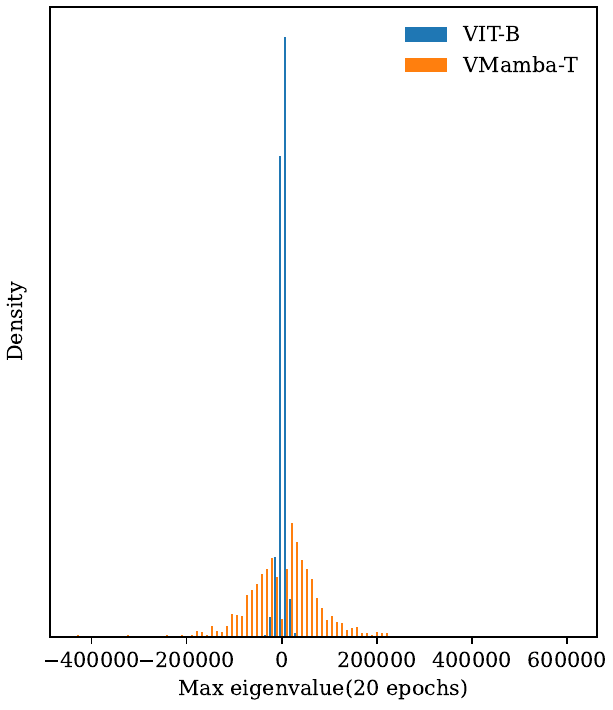}  
    \caption{$20$-th epoch.}  
\end{subfigure}  
\hfill  
\begin{subfigure}{0.30\textwidth}  
    \includegraphics[width=\linewidth]{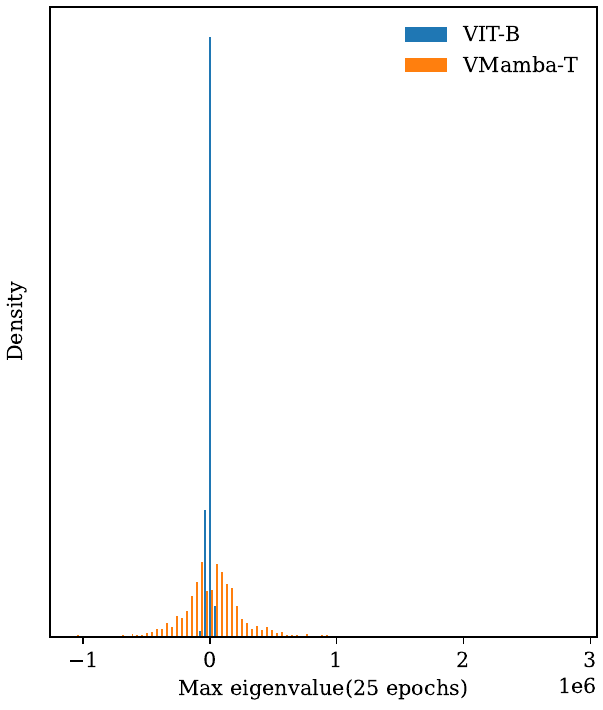}  
    \caption{$25$-th epoch.}  
\end{subfigure}  
\caption{Hessian max eigenvalue spectra.}  
\label{fig:Hessian}  
\end{figure}  

From the aggregate data presented in Figure \ref{fig:OOD}, it's evident that Mamba-based models exhibit superior OOD performance and a pronounced shape bias when compared to their counterparts. This shape bias, indicative of a preference for recognizing the shape of objects over texture, more closely mirrors the image recognition capabilities inherent to human vision \cite{geirhos2021partial}. Such alignment with human visual processing underscores the potential of Mamba-based models in applications requiring nuanced visual understanding.

The more granular insights provided in Figure \ref{fig:OOD_details} further substantiate the dominance of Mamba-based models over those based on ViT architecture. Notably, in conditions where contrast is heightened or a high-pass filter is applied—scenarios, Mamba-based models not only outperform ViT-based models but also surpass human capabilities. On the one hand, both ViTs and human vision exhibit a pronounced bias towards low-frequency components of visual data, as highlighted by \cite{park2022vision}. This predisposition renders them less effective in environments where these components are minimized or absent, such as in the presence of a high-pass filter. On the other, the hiddens of the state space model or Mamba are the coefficients of orthogonal polynomials \cite{gu2020hippo}, and thus the frequency-bias is less evident compared with ViT. 

\subsection{Hessians and Training Landscape}
Hessian spectra reflect the training landscape of the models and a desirable loss landscape is characterized by its flatness and convexity. The Hessian eigenvalues serve as indicators of these characteristics, where the magnitude of the eigenvalues reflects the sharpness of the landscape, and the presence of negative Hessian eigenvalues denotes non-convexity.

We follow \cite{park2022vision} to conduct the analysis. We utilize $3000$ samples with a batch size of $15$. For each batch, we compute the top-5 Hessian eigenvalue spectra, the results of which are depicted in Fig. \ref{fig:Hessian}. The visualization reveals that VMamba models exhibit a higher number of negative eigenvalues compared to ViT models, indicating a more non-convex nature. Furthermore, Mamba models display a greater number of eigenvalues with large magnitudes, suggesting that their loss landscapes are sharper.

\small
\bibliographystyle{nips}
\bibliography{refs}

%%%%%%%%%%%%%%%%%%%%%%%%%%%%%%%%%%%%%%%%%%%%%%%%%%%%%%%%%%%%

\end{document}